# A Novel Just-Noticeable-Difference-based Saliency-Channel Attention Residual Network for Full-Reference Image Quality Predictions

Soomin Seo[†], Sehwan Ki[†], Munchurl Kim[*], *Senior Member, IEEE*

*Abstract*— Recently, due to the strength of deep convolutional neural networks (CNN), many CNN-based image quality assessment (IQA) models have been studied. However, previous CNN-based IQA models likely have yet to utilize the characteristics of the human visual system (HVS) fully for IQA problems when they simply entrust everything to the CNN, expecting it to learn from a training dataset. Therefore, the performance capabilities of such deep-learning-based methods are somewhat saturated. However, in this paper, we propose a novel saliency-channel attention residual network based on the just-noticeable-difference (JND) concept for full-reference image quality assessments (FR-IQA). It is referred to as JND-SalCAR and shows significant improvements in large IQA datasets with various types of distortion. The proposed JND-SalCAR effectively learns how to incorporate human psychophysical characteristics, such as visual saliency and JND, into image quality predictions. In the proposed network, a SalCAR block is devised so that perceptually important features can be extracted with the help of saliency-based spatial attention and channel attention schemes. In addition, a saliency map serves as a guideline for predicting a patch weight map in order to afford stable training of end-to-end optimization for the JND-SalCAR. To the best of our knowledge, our work presents the first HVS-inspired trainable FR-IQA network that considers both visual saliency and the JND characteristics of the HVS. When the visual saliency map and the JND probability map are explicitly given as priors, they can be usefully combined to predict IQA scores rated by humans more precisely, eventually leading to performance improvements and faster convergence. The experimental results show that the proposed JND-SalCAR significantly outperforms all recent state-of-the-art FR-IQA methods on large IQA datasets in terms of the Spearman rank order coefficient (SRCC) and the Pearson linear correlation coefficient (PLCC).

## I. Introduction

RECENTLY, perceptual image quality on smartphones and television displays has become a very important factor that determines superiority among competitive products [68]. Therefore, many companies are studying various image enhancement methods to bolster the competitiveness of their products. However, because human observers are the ultimate consumers of images, the image quality enhancement methods developed thus far often entail a substantial amount of subjective quality verifications, which are cumbersome and time-consuming. Therefore, simple and reliable objective image quality assessment (IQA) is indispensable.

The most commonly used metrics for measuring image quality levels include the simple peak signal-to-noise ratio (PSNR) and the mean square error (MSE). However, it is well known that the PSNR and MSE are not highly correlated with the perceived quality by the human visual system (HVS) [50, 51]. In order to design an accurate IQA model, it is essential to reflect human visual perception characteristics for image quality.

Based on these observations, many computational model-based IQA methods have been proposed by psychophysical experiments [15, 16, 17, 18, 19, 20]. However, such models tend to have high computational complexity and relatively low prediction accuracy levels for various types of distortion.

Recently, deep convolutional neural networks (CNNs) have shown overwhelming performance when applied to most image classification and segmentation problems. Based on these successes, learning-based IQA models have been intensively studied [1, 2, 21, 22, 23, 24, 28, 49]. In the beginning, such models sought directly to predict visual quality scores for single inputs [28]. Currently, CNN-based IQA models are trained to learn visual sensitivity maps, which can be used for weighted pooling for each pixel or patch in the image to predict the image quality score [1, 2, 23]. Visual sensitivity maps were simply learned from the score distributions of the IQA datasets. However, such methods do not take into account the characteristics of the HVS when learning the visual sensitivity map for predictions of visual quality scores on degraded input images. Thus, when HVS-based models (e.g., just-noticeable-difference (JND) [6, 8] and saliency models [7]) are incorporated as inputs to CNN-based IQA models, it is expected that more precise predictions of image quality scores can be attained, as opposed to entrusting all tasks to the CNN.

Inspired by these observations, our deep-learning-based full-reference image quality assessment (FR-IQA) network is designed by incorporating certain perception characteristics of

This research was supported by the MSIT (Ministry of Science and ICT), Korea, under the ITRC (Information Technology Research Center) support program (IITP-2020-2016-0-00288) supervised by the IITP (Institute for Information & communications Technology Promotion).
† Both authors equally contributed to this work.
S. Seo, S. Ki, and M. Kim are with the School of Electrical Engineering, Korea Advanced Institute of Science and Technology, Daejeon 34141, Korea (e-mail: ssm9462, shki@kaist.ac.kr, *corresponding author: mkimee@kaist.ac.kr).

the HVS, in this case visual saliency and the JND. Our contributions can be summarized as follows:

1. We present the first deep-learning-based approach using HVS-based psychophysical models to predict visual scores for degraded images. The two types of input with visual saliency and JND properties are effectively incorporated into an end-to-end trainable network that is dedicatedly designed especially to handle the two input types and to learn the HVS-like perceived visual quality effectively.
2. We propose a novel SalCAR block that extracts perceptually important features using saliency-based spatial attention and channel attention for image quality predictions.
3. The visual saliency map is also used as a guideline for predicting the patch weight map in order to afford stable training of end-to-end optimization for the proposed JND-SalCAR.
4. The JND-SalCAR shows thus far the best prediction accuracy performance in terms of SRCC and PLCC.

This paper is organized as follows. Section II reviews previous works related to the human visual system and various deep-learning-based IQA models in detail. In Section III, we introduce a new JND-based saliency-channel attention residual network, which effectively incorporates HVS-related characteristics. Section IV presents the experimental results of the proposed method in comparison with those of other state-of-the-art FR-IQA models. Section V concludes our JND-SalCAR work.

## II. RELATED WORKS

### A. Human Visual System (HVS)

When the HVS perceives a scene, objects or regions in the scene are not perceived with the same relevance. The characteristics of the HVS for image quality perception can be divided into four categories: the contrast sensitivity function (CSF), luminance masking (LM), contrast masking (CM), and foveated masking (FM). The spatial CSF indicates that the sensitivity of HVS depends on the spatial frequency values [4], indicating that the HVS operates as a band-pass filter by degrading the sensitivity to relatively low- and high-frequency signals. The LM effect refers to how the sensitivity of the HVS is influenced by the degree of background luminance. It has been proved by thorough experiments that the HVS is more sensitive in mid-luminance regions than in relatively dark or bright regions [5]. The CM effect refers to how the sensitivity of the HVS depends on the texture complexity of the background. In background regions with complicated textures, the HVS becomes more insensitive compared to how it operates in homogeneous regions [5]. The FM effect implies that the sensitivity of the HVS is affected by retinal eccentricity from the attention point of the eyes. Distortions that are far from the focus regions are not easily noticeable [6].

JND modeling, which refers to the minimum visibility threshold of the HVS, effectively takes into account these human visual characteristics [6, 8] including CSF, LM, CM and FM. Therefore, the JND model can be used as an important feature in the research related to perceptual quality (e.g., perceptual video coding [8], image quality assessments [26]).

Saliency, referring to the eye-focused region in an image, is also modeled by considering the features of the HVS [27, 73, 74]. It is useful for finding perceptually semantic objects throughout the image. Because the distortions of semantic objects have greater impacts, the saliency map can be used for blur analyses [30] and image quality assessments [16].

The free-energy principle [61, 62] has been widely studied in brain theory and neuroscience, which is introduced to quantize the perception, action, and learning in the human brain. The free-energy principle mainly models the brain activities when perceiving and understanding the visual scenes, which can help further understand the HVS and inspire the study of new IQA methods [62, 63, 64, 65, 66, 67].

### B. Traditional IQA approaches

During the last few decades, much study has been carried out on subjective and objective image/video quality assessment, as reviewed in [25, 68, 69, 70]. Depending on the existence of a reference image for quality assessment of a distorted image, IQA methods can be classified into three categories: full-reference (FR), reduced-reference (RR), and no-reference (NR) IQA methods.

The FR-IQA methods fully utilize the undistorted (original) reference images to measure the qualities of their distorted images. The most common and widely used FR-IQA methods are the mean squared error (MSE) and peak signal-to-noise ratio (PSNR). However, they are poorly correlated with subjective judgments on some distortions [51]. To overcome this problem, Wang *et al*. [25] proposed a structural similarity index (SSIM) based on the fact that HVS is highly optimized to extract structural information from images. The SSIM was extended to multiple scales to the MS-SSIM [14], and its framework of pooling complementary feature similarity maps has served as inspirations for other FR-IQA methods employing different features such as the FSIM [18], GMSD [15], SR-SIM [43], and PSIM [72].

Compared to the FR-IQA methods, NR-IQA methods can predict the image quality without a reference image. Most traditional NR-IQA methods can be classified into Natural Scene Statistics (NSS) methods and learning-based methods. In NSS-based models, blind/referenceless image spatial quality evaluator (BRISQUE) [56] utilizes an asymmetrically generalized Gaussian distribution to model the images in the spatial domain. For more sophisticated modeling, natural image quality evaluator (NIQE) [57] extracts features based on a multivariate Gaussian model and relates them to perceived quality in an unsupervised manner. In learning-based NR-IQA methods, CORNIA [58] is one of the first purely data-driven NR-IQA methods that combines various features for regression. Its codebook is constructed by k-means clustering of luminance and contrast normalized image patches. Recently, Xiongkuo *et al*. [59] proposed a blind PRI-based metric (BPRI) that utilizes a new reference called pseudo-reference image (PRI) generated from the distorted image. Also, they proposed a blind multi PRI-based metric (BMPRI) [60] via distortion aggravation.

*C. Deep-learning-based IQA*

Since the great success of deep learning in many computer vision problems, there have been a few attempts to utilize deep learning for IQA problems. Kang *et al.* [28] was the first to apply a CNN to no-reference (NR) IQA without the use of any handcrafted features. Due to the patch-based training schemes of CNNs, there should be a ground truth score of the image quality for each patch. However, most conventional IQA datasets contain only one visual quality score for each image, representing the overall image quality. Due to this difficulty, Kang's method considered all patches of an image to have the quality scores identical to that of the whole image. In order to overcome this limitation, Kim *et al.* [23] proposed a CNN-based FR-IQA method in which the network predicts the sensitivity map for weighted pooling by feeding an error map of the input image into the network. Bosse *et al.* [2] proposed both FR-IQA and NR-IQA networks that are trained to predict patch-wise image quality scores and patch-wise weights separately via end-to-end training. Weighted pooling is then conducted at the final stage of the network to predict a single quality score for the input image overall.

In addition, consistency of the rank order of the quality scores between the predicted scores and ground truth scores is more important than simply regressing the score values in IQA problems. Liu *et al.* [32] use the rank order of quality scores as a data augmentation technique. Because data with rank orders for various levels of degradation can easily be generated, they successfully trained a large-sized network with plenty of augmented data with rank orders. Prashnani *et al.* [24] proposed a pairwise learning framework for IQA, referred to as PieAPP. They showed that pairwise learning has a great impact on accurate image quality predictions, with their model achieving the highest prediction accuracy among recent IQA methods. The performances of such deep-learning-based FR-IQA methods have reached a point of saturation given their present high prediction accuracy levels. Instead, CNN-based NR-IQA models [45-48] have been intensively studied given that predicting the quality score of a distorted image without access to its reference image is a more challenging problem to solve. For a thorough survey on modern IQA development, please refer to [68, 70, 71].

### III. PROPOSED METHOD

*A. Overview*

In Sections I and II, we discuss how incorporating HVS-related features can boost the accuracy of IQA score prediction. As mentioned in Section II-A, JND is one of the HVS-related visual perception characteristics, which is the minimum visibility threshold of the HVS, where the distortions are hardly recognizable below JND. Also, saliency, referring to the eye-focused region in an image, is an important lead for finding perceptually semantic objects throughout the image. Because the distortions of semantic objects have more significant impacts, the saliency map should be taken into account for precise prediction of quality scores.

Unlike the previous deep-learning-based IQA methods that operate like a black box, our JND-SalCAR network is designed by incorporating the HVS-related features as inputs and guidance when the network learns to process and predict the quality of distorted images. HVS-related features can be a big help in terms of the prediction accuracy when predicting the quality scores that are rated by humans. Fig. 1 shows the architecture of the proposed JND-SalCAR. JND-SalCAR is composed of three subnets in total: a subnet for reference and distorted input images (ImgSubnet); a subnet for saliency map input (SalSubnet); and a subnet for JND probability map input (JndSubnet). ImgSubnet is devised to extract features from reference and distorted images separately. Their feature difference is handed over to the latter part of the network to predict the quality score of the distorted image compared to its reference image. SalSubnet extracts features from the saliency map, which contains very important information that indicates where a person focuses when viewing an image. Because the distortions in salient regions have a greater impact on IQA scores, saliency maps can act as good guidance to tell the network where to focus on the distorted image. Unlike the pixel-wise feature difference between reference and distorted images from ImgSubnet, JndSubnet gives an important clue where a human is likely to perceive a larger error, yielding a perceptual error map for the given distorted image. For the network to map the features from an image domain to a score domain, the spatial sizes of feature maps are reduced via max-pooling operations.

*B. Architecture*

JND-SalCAR takes four types of input: a distorted image, a reference image, its saliency map, and the corresponding JND probability map. Unlike conventional methods [1, 2, 22, 23, 24, 28] that only use reference and distorted images as inputs, JND-SalCAR utilizes saliency maps and JND probability maps as additional important inputs to extract HVS-based perceptual features, allowing it to more precisely predict the perceptual quality levels of distorted images

The feature-extraction part of the JND-SalCAR is composed of a subnet for reference and distorted input images (ImgSubnet), a subnet for saliency map input (SalSubnet), and a subnet for JND probability map input (JndSubnet).

ImgSubnet is a Siamese network, which constitutes two identical networks sharing their weights when extracting the features from the *i*-th reference input patch $p_i^{Ref}$ and the *i*-th distorted input patch $p_i^{Dst}$. By extracting useful information for predicting the visual quality, the features $F_i^{Ref}$ and $F_i^{Dst}$ from the corresponding input patches $p_i^{Ref}$ and $p_i^{Dst}$ are obtained as outputs of ImgSubnet. $F_i^{Ref}$ and $F_i^{Dst}$ are expressed correspondingly as

$$F_{i\in\{1,2,\cdots,N\}}^{Ref} = \text{ImgSubnet}(p_i^{Ref}, F_i^{Sal-1}, F_i^{Sal-2}), \quad (1)$$

$$F_{i\in\{1,2,\cdots,N\}}^{Dst} = \text{ImgSubnet}(p_i^{Dst}, F_i^{Sal-1}, F_i^{Sal-2}), \quad (2)$$

where *N* is the total number of patches for an input image. As explained above, ImgSubnet is a shared net for both $F_i^{Ref}$ and $F_i^{Dst}$. In (1) and (2), $F_i^{Sal-1}$ and $F_i^{Sal-2}$ are the intermediate feature maps from SalSubnet, as explained below.

SalSubnet, which receives the *i*-th saliency map patch $p_i^{Sal}$ as input, consists of four convolution layers where $F_i^{Sal-1}$ and $F_i^{Sal-2}$ are fused via concatenation and are inputted to the first and second SalCAR blocks of ImgSubnet and JndSubnet. $F_i^{Sal-1}$ and $F_i^{Sal-2}$ are correspondingly the output feature maps after the first and second max-pooling layers of SalSubnet, expressed as

$$F_{i\in\{1,2,\cdots,N\}}^{Sal-1} = MaxPool(Conv3 \circ Conv3(p_i^{Sal})), \quad (3)$$

$$F_{i\in\{1,2,\cdots,N\}}^{Sal-2} = MaxPool(Conv3 \circ Conv3(F_i^{Sal-1})), \quad (4)$$

where *MaxPool* is the max-pooling layer and *Conv3* is the convolution layer with $3 \times 3$ kernels.

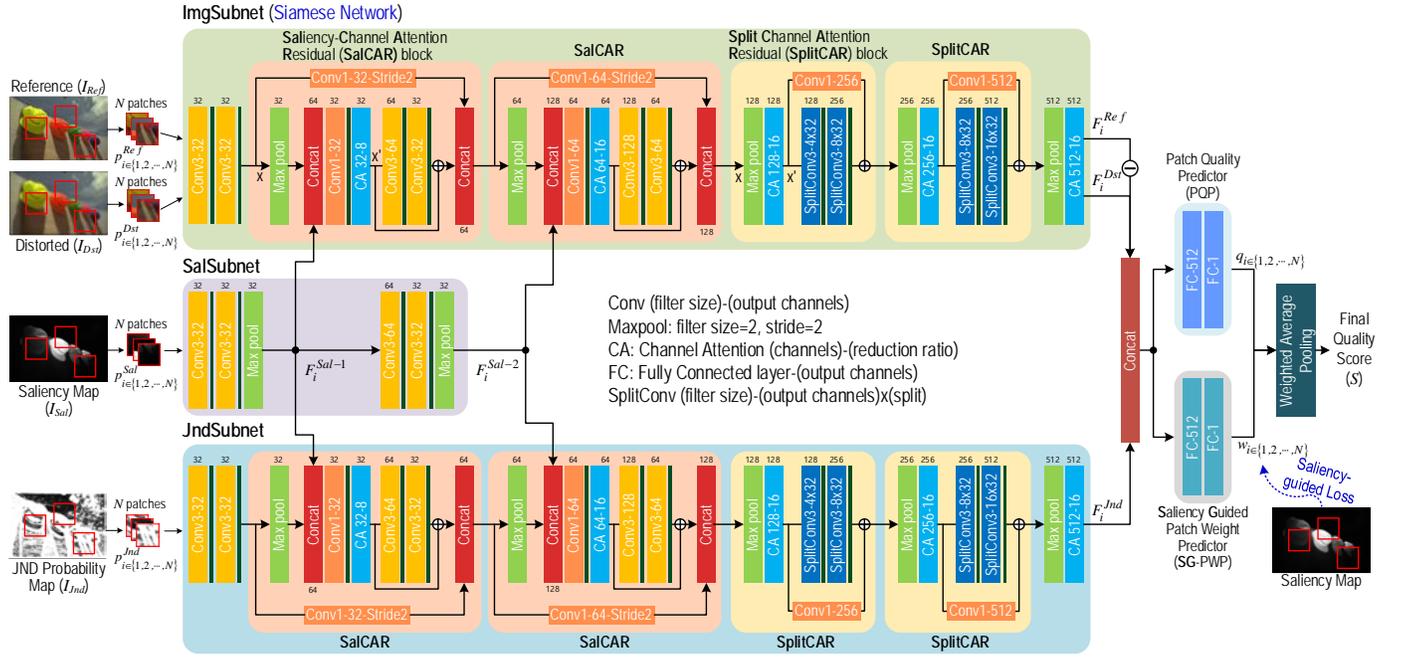

Fig. 1. Architecture of JND-SalCAR. Features are extracted from reference patches, distorted patches, and JND probability map patches by cascaded SalCAR blocks and SplitCAR blocks. The two extracted features, $F_i^{Ref}$ and $F_i^{Dst}$, are subtracted and concatenated with $F_i^{JND}$. $F_i^{Ref}$ and $F_i^{Dst}$ are extracted using a Siamese network, indicated by ImgSubnet in Fig. 1. The concatenated feature vector is input to the saliency guided patch weight predictor and the patch quality predictor, and weighted averaging pooling is used to gather all of the scores from the patches in one image. The thin bars after the convolution layers indicate leaky ReLU operations.

JndSubnet takes the *i*-th JND probability map patch $p_i^{Jnd}$ as input and uses the same architecture as ImgSubnet, but its weights are not shared with those of ImgSubnet.

$$F_{i\in\{1,2,\cdots,N\}}^{Jnd} = \text{JndSubnet}(p_i^{Jnd}, F_i^{Sal-1}, F_i^{Sal-2}) \quad (5)$$

By utilizing $F_i^{Sal-1}$ and $F_i^{Sal-2}$ in (1), (2) and (5), both ImgSubnet and JndSubnet can be assisted by the saliency context to extract effective features from visually focused information.

The feature-extraction part of JND-SalCAR finally yields the feature vectors, $F_i^{Ref}$, $F_i^{Dst}$ and, $F_i^{JND}$, where the ImgSubnet yields $F_i^{Ref}$ and $F_i^{Dst}$ and JndSubnet produces $F_i^{JND}$.

Because the difference between the two feature vectors $F_i^{Ref}$ and $F_i^{Dst}$ is a very important clue when predicting the perceptual quality scores of degraded images [1], we construct a large feature vector by concatenating the feature vector difference ($F_i^{Ref}$ - $F_i^{Dst}$) to $F_i^{JND}$, as indicated by the red box with the operation *Concat* in Fig. 1. The augmented vector after concatenation is then utilized as the input to both the saliency-guided patch weight predictor (SG-PWP) and the patch quality predictor (PQP), as shown in Fig. 1. The SG-PWP and PQP have an identical structure consisting of two fully connected layers of 512 outputs and one output. The outputs of SG-PWP and PQP can be interpreted as the estimated patch weights ($w_i$) and the patch quality scores ($q_i$), respectively. $w_i$ and $q_i$ are given by

$$w_{i\in\{1,2,\cdots,N\}} = \text{SG-PWP}(Concat(F_i^{Ref} - F_i^{Dst}, F_i^{Jnd})) \quad (6)$$

$$q_{i\in\{1,2,\cdots,N\}} = \text{PQP}(Concat(F_i^{Ref} - F_i^{Dst}, F_i^{Jnd})) \quad (7)$$

After obtaining $w_i$ and $q_i$ for each patch in a degraded input image, the weighted average-pooling module finally produces one final predicted quality score $S$ for the entire degraded input image using the weighted average pooling of $w_i$ and $q_i$. The final predicted quality score $S$ is calculated as shown below.

$$S = \sum_{i=1}^{N} w_i \cdot q_i \quad (8)$$

### C. JND Probability

In FR-IQA, it is important to measure the difference between an original image and its distorted image. Kim *et al.* [23] used the difference between the edge map of an original reference image and that of a distorted image for IQA. In another study [2], the difference between the feature map of an original image and that of a distorted image is used as input for an image quality predictor. Intuitively, the HVS-based perceptual difference value rather than the numerical difference value can be used to predict the image quality score more accurately. To do this, we use a distortion detection probability map, referred to as the JND probability map, to extract effective features. We used the DCT-based JND model [8] to calculate the JND probability maps for distorted images and then feed them into JndSubnet to predict the image quality.

With the input JND probability map, the JND-SalCAR is trained to learn different sensitivities of the HVS for various distortion types and amounts. Figs. 2-(c) and (d) respectively show the squared image difference (SID) and the JND probability map between the original image in Fig. 2-(a) and the distorted image corrupted by additive white Gaussian noise (AWGN) in Fig. 2-(b). Figs. 2-(g) and (h) show the SID map and the JND probability map between the original image in Fig. 2-(a) and the distorted image corrupted by Gaussian blur (GB) in Fig. 2-(f). As shown in Fig. 2-(b), AWGN distortions are easily perceived in the homogeneous regions due to the CM effects of the HVS [5]. This can be confirmed in Fig. 2-(d), which indicates that higher JND probability values are observed in the homogeneous regions (brighter regions), such as the background area. On the other hand, the SID map in Fig. 2-(c) simply shows the AWGN distortion that uniformly appears but which does not reflect the HVS's visual perception characteristic. On the other hand, GB distortion tends to be easily perceived in complex texture regions. As shown in Fig. 2-(h), higher distortion detection probability values are observed in complex-textured regions, such as the butterfly and the flowers (brighter regions). Thus, JND probability maps can be effectively incorporated to reflect the HVS's distortion perception characteristics well in predictions of subjective IQA scores than using numerical error maps.

*D. SalCAR Block*

For a deep-learning-based IQA method, the performance and the representational power greatly depend on the network's capability to capture features that make greater contributions and to focus on learning features that are more important. To achieve this, we design a saliency-guided channel attention residual (SalCAR) block that captures both spatially and channel-wise important information with the help of a saliency map and a channel attention (CA) unit. As shown in Fig. 1, each SalCAR block is composed of a max-pooling layer, concatenation of saliency map features and the output of the max-pooling layer, three convolution layers, a channel attention unit and two different skip connection types.

1) *Saliency:* Saliency plays an important role in IQA, as distorted image regions with higher levels of visual attention tend to affect the visual quality of the entire image with a stronger influence. As another aspect of utilizing the HVS characteristics, the saliency map is utilized as input to the proposed JND-SalCAR network. In order to calculate the visual saliency maps, we adopt an approach based on the minimum barrier detection concept [7]. Fig. 2-(e) shows a saliency map of the original image in Fig. 2-(a).

As explained previously and as shown in Fig. 1, each patch of an input saliency map is fed into SalSubnet and its feature outputs obtained after the first and second max-pooling layers are fused via concatenation into the first and second SalCAR blocks respectively. By feeding this saliency context into SalCAR blocks, the focus can be on spatial feature regions that attract greater levels of visual attention. After each feature concatenation, 1×1 convolution is utilized in order to modulate the number of channels in ImgSubnet and JndSubnet, as is done before feature concatenation.

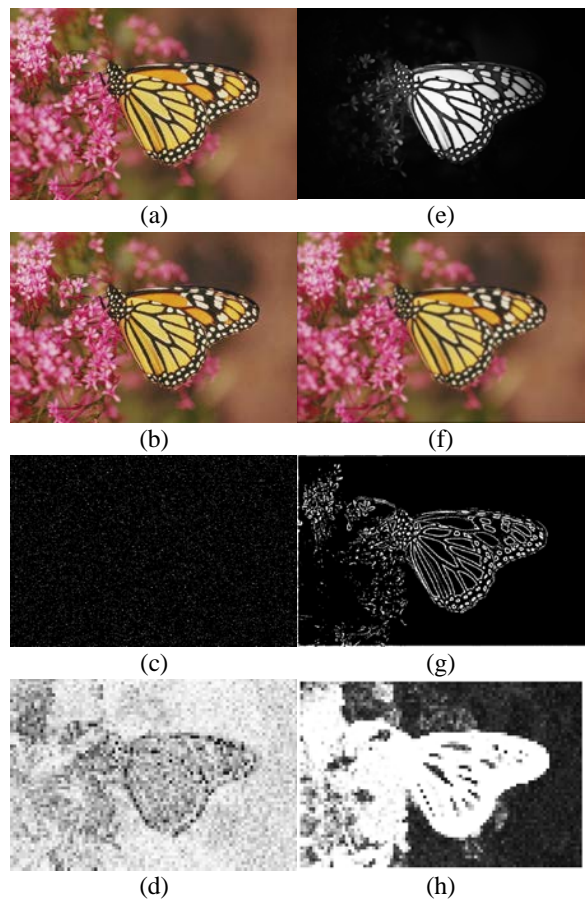

Fig. 2. Examples of the saliency map and the JND probability map. (a) is original image and (e) is the saliency map of (a). (b) is WN image of (a), and (c), (d) are the squared image difference (SID) map and the JND probability map for (b). (f) is GB image and (g), (h) are the SID map and the JND probability map for (f).

It should be noted that because the initial patch is 32 × 32 in size and given that the feature map size is reduced to 8 × 8 after the second max-pooling layer, the effect of the saliency context is diminished for such a small 8 × 8-sized feature map. Therefore, SalSubnet only incorporates the saliency context immediately after the first two max-pooling layers.

2) *Channel Attention (CA):* Inspired by its superior performance in image captioning [9] and super-resolution [10], channel attention is adopted in JND-SalCAR. Given the input feature map to the CA unit, each channel is pooled to its average value via global average pooling (GP). After the GP operation, channel downscaling is done via 1×1 convolution with a reduction ratio, $r$, followed by ReLU activation. The reduced feature vector is then increased back to the original amount with the ratio $r$, using a channel-upscaling layer. The cha in order to determine the channel attention factors which are later adapted by a sigmoid function. Then, each channel of the input feature map is rescaled by its corresponding channel attention factor, which results in a refined feature map. It has precisely the same shape as the original feature maps but contains differently focused feature maps from a channel-wise perspective, which allows more informative feature maps to affect the prediction of subjective IQA scores with a greater influence.

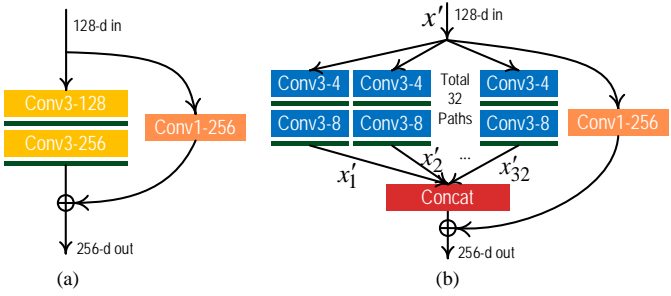

Fig. 3. (a) A block with plain convolution layers. (b) A block with a split convolution operation. The thin bars indicate leaky ReLU operations.

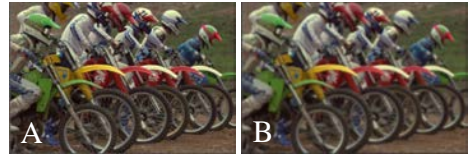

Fig. 4. Illustration on the effect of the rank order for IQA

*3) Skip Connection:* As the size of the network grows, it becomes more difficult to train the network appropriately because the information flow becomes weaker and the gradient vanishing problem arises. To tackle this issue, two different types of skip connections are applied to the SalCAR block. The first skip connection in each SalCAR block is a simple parameter-free identity shortcut [33] around two stacked 3×3 convolution layers. The second skip connection incorporates the concatenation of the features after the first skip connection and another feature computed from the initial feature map of the SalCAR block by 1×1 convolution with a stride of 2 in order to match the size of the feature map. With these two different skip connection types, SalCAR blocks improve the information flow throughout the proposed JND-SalCAR network. Furthermore, the skip connection recovers information that may have been lost during the channel attention and saliency-guided feature extraction process.

For input $x$, each SalCAR block in Fig. 1 performs a series of max-pooling, concatenation, convolution and channel attention operations, by which the output is produced, expressed as shown below.

$$SalCAR(x) = Concat\{Conv3 \circ Conv3(x') + x', Conv1(x)\} \quad (9)$$

Here, $x' = CA(Conv1(Concat\{MaxPool(x), F_{sal}^1\}))$, obtained via a channel attention module, is denoted as *CA*.

*E. SplitCAR Block*

The output of the second SalCAR block is further fed into the split convolution channel attention residual (SplitCAR) block, as shown in Fig. 1. As noted above, the saliency context is no longer used in SplitCAR blocks where the size of the feature map is too small to make good use of the saliency information.

In the proposed JND-SalCAR network, the size of the feature map decreases and the number of channels increases as the network becomes deeper. This approach encounters a problem in that the number of parameters increases exponentially. Inspired by the ResNeXt [34] design, we propose a SplitCAR block using split convolution (Fig. 3), where a convolution operation is split into several branches and aggregated later by concatenation. This approach can greatly reduce the number of parameters while maintaining the performance of the network.

The output of the SplitCAR block in Fig. 1 and Fig. 3-(b) is then given by

$$SplitCAR(x) = Concat\{x'_1, x'_2, \cdots, x'_{32}\} + Conv1(x'), \quad (10)$$

where Conv1 indicates convolution with 1×1 filters. In addition, $x'$ and $x'_i$ in (11) are intermediate outputs of SplitCAR, given by

$$x' = CA(MaxPool(x)), \quad (11)$$

$$x'_i = Conv3(Conv3(x')), \quad (12)$$

where $i$ is an index of split convolutions, running from 1 to 32, as shown in Fig. 3-(b).

*F. Saliency-Guided Loss*

The weighted-average-pooling method [2] was proposed to predict the entire image quality score using the predicted patch quality scores and patch weights. However, separate networks are trained without any ground truth to predict the patch weights and patch quality scores. Therefore, it is difficult to conclude that both networks actually play their predefined roles. Specifically, it is not appropriate to determine the patch weights only with the input patch without considering the overall image information.

Thus, we newly propose a saliency-guided patch weight predictor (SG-PWP) with saliency-guided loss. A visual saliency map can serve as a good guideline for the SG-PWP to learn which patch should be perceived with greater importance considering the entire image. The normalized $l$-th patch weight $\hat{w}_l$ is given by

$$\hat{w}_l = w_l \Big/ \sum_{i=1}^{N} w_i, \quad (13)$$

where $w_i$ is the $i$-th estimated patch weight and $N$ is the total number of patches randomly extracted from a training image. Similarly, visual saliency significance $v_l$ for the $l$-th patch ($p_l$) in a visual saliency map is defined as

$$v_l = \sum_{(m,n) \in R_l} I_s(m,n) \Big/ \sum_{(i,j)} I_s(i,j), \quad (14)$$

where $I_s(m,n)$ is the visual saliency value at location $(m, n)$ in region $R_l$ of $p_l$, while $I_s(i,j)$ is the visual saliency value at location $(i, j)$ in the total region of the visual saliency map $I_s$. $v_l$ in (14) is proportional to the total sum of the visual saliency values within its patch. Based on (13) and (14), the saliency-guided loss term for the $k$-th distorted image is defined as follows:

$$L_{Sal}(I_k; \theta) = \frac{1}{N} \sum_{i=1}^{N} |\hat{w}_i - v_i| \quad (15)$$

### G. Rank Loss

At this point, we consider a situation in which two distorted images are presented to human subjects in a subjective quality assessment. Sometimes it is challenging for individuals directly to rate quality scores for two images, and the rated scores can even depend heavily on the individual. However, it is relatively easy to tell which image is of a better quality between the two. In addition, consistency of the rank order of the quality scores between the predicted scores and the ground truth scores is more important than simply regressing the score values in IQA problems.

Fig. 4 illustrates an example of rank order significance for an IQA problem. In Fig. 4, Image A has a higher MOS value than Image B. IQA methods 1 and 2 produce the same MAE value between the ground truth MOS and the predicted MOS (pMOS) values. Therefore, in terms of regression, such a case is considered to have produced the same performance outcomes. However, in terms of rank statistics, IQA method 2 yielded pMOS values in the reverse order compared to the ground truth and IQA method 1. Therefore, it is essential to take into account the rank order of pMOS values in IQA.

Motivated by this, the concept of "rank loss" was proposed and studied to reflect the rank order among degraded images in IQA methods. In one study [31], the quality preference for each given image pair was explored, but the loss is only considered for preference direction without taking into account the preference score difference. In another study [32], the rank loss was used in a limited manner for each pair of two distorted images for the same reference image with the same distortion type but at different amounts (e.g., a pair of two distorted *Lena* images, $I_A^d$ and $I_B^d$, with Gaussian blur of $\sigma = 1$ and $\sigma = 4$). Furthermore, the rank loss was only used during the pre-training of the network. In PieAPP [24], the network was trained to learn the preference probability for each image pair with different distortions.

We incorporate the rank loss proposed in earlier work [31] into the training of JND-SalCAR, which penalizes the network when the rank of the output scores does not agree with the rank of the ground truth scores regardless of their distortion types. Additionally, we apply a joint optimization framework that considers both the predictions of the quality scores and their rank orders for deep-learning-based IQA.

Given a set of distorted images $I_n^{Dst}$, $n = 1, ..., N_B$, where $N_B$ is the batch size, as the network input, let $f(I_n^{Dst}; \theta)$ be the final predicted score of JND-SalCAR, where $f$ is the overall network function and $\theta$ denotes the network parameters. The ground truth score for a distorted image $I_n^{Dst}$ is denoted by $s_n$. The pairwise rank loss between two distorted images, $I_x^{Dst}$ and $I_y^{Dst}$, can then be computed as

$$L(I_x^{Dst}, I_y^{Dst}; \theta) = \max\left(0, \frac{-(s_x - s_y) \cdot (f(I_x^{Dst}; \theta) - f(I_y^{Dst}; \theta))}{|s_x - s_y| + \varepsilon}\right), \quad (16)$$

where $\varepsilon$ is a small stability term. When the ranks of the predicted scores and ground truth scores match, $L(I_x^{Dst}, I_y^{Dst}; \theta)$ is always 0; otherwise, it is reduced to the absolute difference of the corresponding pMOS values, as

$$L(I_x^{Dst}, I_y^{Dst}; \theta) = |f(I_x^{Dst}; \theta) - f(I_y^{Dst}; \theta)|. \quad (17)$$

Because the rank loss in (16) can be calculated for each pair of distorted images, we have $_{N_B}C_2$ combinations of losses for a batch size of $N_B$. Therefore, the resulting total rank loss is given as the sum of the six terms by

$$L_{rank} = L(I_1^{Dst}, I_2^{Dst}; \theta) + L(I_1^{Dst}, I_3^{Dst}; \theta) + L(I_1^{Dst}, I_4^{Dst}; \theta) \\ + L(I_2^{Dst}, I_3^{Dst}; \theta) + L(I_2^{Dst}, I_4^{Dst}; \theta) + L(I_3^{Dst}, I_4^{Dst}; \theta). \quad (18)$$

Consequently, the total loss when training JND-SalCAR is composed of three terms: the mean absolute error (MAE) loss, the saliency-guided loss in (15) and the rank loss in (18). The total loss can be evaluated as the weighted sum of the three loss terms, as

$$L_{tot} = \alpha \times L_{MAE} + \beta \times L_{Rank} + \gamma \times L_{Sal}, \quad (19)$$

where the hyper-parameters $\alpha = 1$, $\beta = 10$, $\gamma = 1$ are set empirically based on a series of experiments. These values are the default setting in the following experiments.

This enables the joint optimization of both the predictions of the quality scores and their rank orders, which is differentiated from the previous works that consider the rank order of the distorted images [24, 31, 32].

### H. Training

All images are rescaled to the range of [-0.5, 0.5] from their original range [0, 255]. The input image set includes four different types of images: reference images, distorted images, saliency maps, and JND probability maps. The patches from the four types of images are sampled together at the same locations and then fed into ImgSubnet, SalSubnet and JndSubnet of JND-SalCAR. One training sample is a quaternion that consists of four patches, which are cropped from a reference image, a distorted image, a saliency map and a JND probability map at the same location. The batch size is set to 4, and the patch size is 32×32. Each training image is represented by 32 randomly sampled patches for data augmentation. For model training, we use the Adam optimizer [29] with an initial learning rate of $10^{-4}$. During validation, we use all non-overlapping 32×32-sized patches of a validation input image to improve the accuracy of the image quality predictions. Thus, the number of extracted patches of a validation input image differs depending on the validation input image size.

We implemented the proposed networks with the TensorFlow[TM] [35] framework and trained them using NVIDIA Titan Xp[TM] GPUs. It takes 12 hours to train the network for 1,000 epochs, and the test runtime per image of size 512×384 takes 23 ms.

## IV. EXPERIMENTAL RESULTS

### A. Dataset

We assess the effectiveness of the proposed JND-SalCAR on four different image quality datasets: LIVE [3], CSIQ [11],

TABLE I. PERFORMANCE COMPARISON ON FOUR DIFFERENT IQA DATASETS

| METHOD | LIVE [1] | | | CSIQ [2] | | | TID2008 [3] | | | TID2013 [4] | | |
|---|---|---|---|---|---|---|---|---|---|---|---|---|
| | SRCC (std) | PLCC (std) | KRCC (std) | SRCC (std) | PLCC (std) | KRCC (std) | SRCC (std) | PLCC (std) | KRCC (std) | SRCC (std) | PLCC (std) | KRCC (std) |
| MAPE | 0.730 | 0.514 | 0.558 | 0.109 | 0.099 | 0.078 | 0.268 | 0.377 | 0.189 | 0.257 | 0.301 | 0.177 |
| MRSE | 0.748 | 0.657 | 0.582 | 0.168 | 0.123 | 0.123 | 0.269 | 0.306 | 0.182 | 0.349 | 0.400 | 0.237 |
| NQM [36] | 0.903 | 0.908 | 0.737 | 0.821 | 0.824 | 0.651 | 0.611 | 0.600 | 0.448 | 0.701 | 0.701 | 0.527 |
| IFC [37] | 0.925 | 0.929 | 0.761 | 0.821 | 0.852 | 0.641 | 0.651 | 0.592 | 0.472 | 0.616 | 0.577 | 0.447 |
| VIF [38] | 0.961 | 0.960 | 0.824 | 0.905 | 0.915 | 0.732 | 0.759 | 0.819 | 0.598 | 0.687 | 0.778 | 0.523 |
| VSNR [39] | 0.868 | 0.909 | 0.718 | 0.819 | 0.810 | 0.629 | 0.713 | 0.690 | 0.527 | 0.739 | 0.734 | 0.552 |
| MAD [40] | 0.960 | 0.952 | 0.838 | 0.946 | 0.948 | 0.800 | 0.824 | 0.821 | 0.639 | 0.816 | 0.835 | 0.633 |
| RFSIM [41] | 0.955 | 0.961 | 0.828 | 0.928 | 0.917 | 0.769 | 0.865 | 0.868 | 0.678 | 0.811 | 0.843 | 0.624 |
| GSM [42] | 0.944 | 0.948 | 0.806 | 0.912 | 0.905 | 0.737 | 0.792 | 0.788 | 0.590 | 0.802 | 0.828 | 0.609 |
| SR-SIM [43] | 0.957 | 0.947 | 0.826 | 0.932 | 0.929 | 0.776 | 0.890 | 0.889 | 0.715 | 0.808 | 0.873 | 0.644 |
| MDSI [44] | 0.961 | 0.972 | 0.831 | 0.957 | 0.957 | 0.820 | 0.918 | 0.916 | 0.748 | 0.889 | 0.912 | 0.715 |
| MAE | 0.880 | 0.860 | 0.714 | 0.836 | 0.820 | 0.645 | 0.360 | 0.393 | 0.246 | 0.573 | 0.588 | 0.405 |
| RMSE | 0.899 | 0.849 | 0.737 | 0.861 | 0.866 | 0.668 | 0.564 | 0.560 | 0.402 | 0.709 | 0.698 | 0.516 |
| SSIM [25] | 0.948 | 0.942 | 0.812 | 0.861 | 0.853 | 0.683 | 0.775 | 0.777 | 0.578 | 0.753 | 0.805 | 0.574 |
| MS-SSIM [14] | 0.956 | 0.946 | 0.829 | 0.907 | 0.897 | 0.738 | 0.844 | 0.841 | 0.647 | 0.797 | 0.854 | 0.623 |
| GMSD [15] | 0.957 | 0.955 | 0.829 | 0.957 | 0.958 | 0.817 | 0.884 | 0.877 | 0.703 | 0.804 | 0.865 | 0.635 |
| VSI [16] | 0.948 | 0.946 | 0.807 | 0.944 | 0.936 | 0.792 | 0.902 | 0.885 | 0.720 | 0.900 | 0.907 | 0.725 |
| PSNR-HMA [17] | 0.946 | 0.956 | 0.809 | 0.912 | 0.913 | 0.744 | 0.838 | 0.827 | 0.653 | 0.814 | 0.837 | 0.635 |
| FSIMc [18] | 0.963 | 0.953 | 0.846 | 0.928 | 0.922 | 0.770 | 0.884 | 0.880 | 0.702 | 0.853 | 0.885 | 0.673 |
| SFF [19] | 0.970 | 0.956 | 0.860 | 0.962 | 0.966 | 0.830 | 0.883 | 0.891 | 0.699 | 0.857 | 0.880 | 0.668 |
| SCQI [20] | 0.946 | 0.940 | 0.803 | 0.949 | 0.942 | 0.802 | 0.908 | 0.895 | 0.733 | 0.908 | 0.917 | 0.739 |
| PSIM [72] | 0.961 | 0.951 | 0.839 | 0.961 | 0.965 | 0.827 | 0.910 | 0.909 | 0.738 | 0.893 | 0.911 | 0.720 |
| DOG-SSIMc [21] | 0.963 | 0.966 | 0.844 | 0.954 | 0.943 | 0.813 | 0.935 | 0.939 | 0.786 | 0.926 | 0.934 | 0.768 |
| Lukin *et al*. [22] | - | - | - | - | - | - | - | - | - | 0.930 | - | 0.770 |
| Kim *et al*. [23] | 0.981 | 0.982 | - | 0.961 | 0.965 | - | 0.947 | 0.951 | - | 0.939 | 0.947 | - |
| Bosse *et al*. [2] | 0.970 | 0.980 | - | - | - | - | - | - | - | 0.940 | 0.946 | - |
| PieAPP [24] | 0.977 | 0.986 | 0.894 | 0.973 | 0.975 | 0.881 | 0.951 | 0.956 | 0.822 | 0.945 | 0.946 | 0.804 |
| **JND-SalCAR** | **0.984** (0.003) | **0.987** (0.002) | **0.899** (0.009) | **0.976** (0.003) | **0.977** (0.003) | 0.868 (0.009) | **0.956** (0.003) | **0.957** (0.004) | 0.821 (0.006) | **0.949** (0.004) | **0.956** (0.004) | **0.812** (0.008) |

TID2008 [12] and TID2013 [13]. The LIVE [3] image quality dataset contains 799 distorted images from 29 reference images. Five different types of distortions are included: JP2K compression (JP2K), JPEG compression (JPEG), additive white Gaussian noise (WN), Gaussian blur (BLUR) and Rayleigh fast fading channel distortion (FF). The CSIQ [11] image quality dataset consists of 866 distorted images from 30 reference images. The images are distorted by six different distortion types: JPEG, JP2K, WN, BLUR, additive pink Gaussian noise (PGN) and global contrast decrements (CTD). The TID2008 [12] image quality dataset contains 1,700 distorted images from 25 reference images. Here, 17 different types of distortions are included. The TID2013 [13] image quality dataset is an extended version of the TID2008 dataset. It has the same set of reference images as TID2008 but a greater variety of distortions, with 24 different distortion types introduced in this case.

All image quality ratings for four different datasets were realigned and normalized to have a range of [0, 9], where a higher value indicates perceptually better quality. Each dataset was randomly divided into three subsets according to their reference images for training, validation, and testing. For the LIVE dataset, 17 reference images are used for training, six images are used for validation, and six are for testing out of 29 reference images in total. The CSIQ dataset is split into 20 training, five validation and five test images. For the TID2008 and TID2013 datasets, 25 reference images are divided into 15 images for training, five images for validation, and five images for testing.

### B. Performance Comparison

JND-SalCAR is trained for 1,000 epochs, where an epoch describes the number of times the network has seen all of the samples in the entire training dataset. The model with the lowest validation loss is chosen as the final model. The performances of the proposed method and various IQA algorithms are evaluated in terms of certain correlation metrics, in this case the Spearman rank order coefficient (SRCC), the Pearson linear correlation coefficient (PLCC) and the Kendall rank correlation coefficient (KRCC). For each correlation metric, values closer to 1 indicate higher performances.

All experiments were repeated five times and the results were averaged for a fair comparison. Each dataset was randomly divided into training, validation and test image sets every time we started the model training process. JND-SalCAR is compared to a total of 27 IQA metrics. Twenty-two non-deep-learning-based IQA methods are included: the mean absolute error (MAE), root mean square error (RMSE), SSIM [25], MS-SSIM [14], GMSD [15], VSI [16], PSNR-HMA [17], FSIMc [18], SFF [19], SCQI [20], NQM [36], IFC [37], VIF [38], VSNR [39], MAD [40], RFSIM [41], GSM [42], SR-SIM [43], MDSI [44], and PSIM [72]. Five deep-learning-based methods are included: DOG-SSIMc [21], Lukin *et al.* [22], Kim *et al.* [23], Bosse *et al.* [2], and PieAPP [24].

TABLE II. PERFORMANCE COMPARISON ON KADID-10K [54] DATASET.

| METHOD | KADID-10K [54] | |
| --- | --- | --- |
| | SRCC | PLCC |
| MAPE | 0.481 | 0.498 |
| MRSE | 0.543 | 0.506 |
| MAD [40] | 0.809 | 0.808 |
| RFSIM [41] | 0.849 | 0.848 |
| GSM [42] | 0.798 | 0.797 |
| SR-SIM [43] | 0.845 | 0.841 |
| MDSI [44] | 0.899 | 0.899 |
| MAE | 0.465 | 0.504 |
| RMSE | 0.698 | 0.712 |
| SSIM [25] | 0.783 | 0.780 |
| MS-SSIM [14] | 0.838 | 0.835 |
| GMSD [15] | 0.856 | 0.854 |
| VSI [16] | 0.893 | 0.893 |
| PSNR-HMA [17] | 0.842 | 0.833 |
| FSIMc [18] | 0.871 | 0.869 |
| SFF [19] | 0.875 | 0.875 |
| SCQI [20] | 0.869 | 0.870 |
| PSIM [72] | 0.890 | 0.890 |
| DeepFL-IQA [75] | 0.936 | 0.938 |
| DualCNN [55] | 0.941 | 0.949 |
| **JND-SalCAR** | **0.959** | **0.960** |

Table I compares the SRCC, PLCC and KRCC performances for the 27 IQA algorithms and the proposed JND-SalCAR with four different datasets. It should be noted in Table I that the SRCC, PLCC, and KRCC values for five deep-learning-based IQA methods come from the corresponding original papers. The highest and the second highest SRCC, PLCC and KRCC values in Table I are shown in boldface and underlined text, respectively.

As indicated in Table I, JND-SalCAR consistently outperforms all of the state-of-the-art IQA methods on all IQA datasets in terms of SRCC and PLCC. Moreover, Table I shows that the proposed method achieves significant gains over the state-of-the-art methods for TID2013, which is the largest IQA dataset with the largest number of distortion types. For the LIVE, CSIQ and TID2008 IQA datasets, although the prediction performances of the state-of-the-art methods have already reached very high accuracy levels for visual quality scores, JND-SalCAR improved on these values compared to highest performing state-of-the-art method, PieAPP [24], which even uses a larger network by five times (18,147K parameters) compared to that used here (3,547K). Accordingly, it can be concluded from these observations that the performance outcomes can be improved with even fewer training parameters when the HVS-based model is used for deep learning training. The superiority of JND-SalCAR is even more apparent in terms of SRCC as it outperforms the other algorithms with a large margin. From this observation, we can infer that the rank loss adequately contributes to a more accurate pMOS prediction by considering the relationships among quality score orders among the rated distorted images.

In the KRCC performance comparison in Table I, JND-SalCAR is shown to outperform all SOTA methods for LIVE and TID2013, though it shows lower performance and comparable performance relative to those of PieAPP [24] for CSIQ and TID2008, respectively. However, it should be noted that TID2013 has the largest dataset (3,000 images with 24 distortion types and 985 evaluators), while CSIQ is a relatively small dataset (866 images with only six distortion types and only 35 evaluators); additionally, TID2008 is a subset (1,700 images with 17 distortion types and 838 evaluators) of TID2013. Because the proposed JND-SalCAR approach significantly outperforms all of the SOTA methods on the largest dataset (TID2013), it can be concluded that JND-SalCAR works reasonably well.

We evaluate JND-SalCAR further on a recently proposed IQA dataset, KADID-10k [54]. The KADID-10k [54] image quality dataset contains 10,125 distorted images from 81 pristine images. Each of the reference images is degraded by 25 distortions at five levels. In this case, 17 different types of distortions are included. Table II compares the conventional and deep-learning-based IQA methods for the KADID-10k image quality dataset. From Table II, it is clearly indicated that JND-SalCAR outperforms all conventional non-deep-learning methods and recently proposed deep-learning-based IQA methods as well.

*C. Ablation study*

To examine the effectiveness of the conjunction of network components used with JND-SalCAR, we conducted a detailed performance analysis of the networks with different combinations of network components. Table III shows the SRCC performances for the different combinations of network components for the TID2013 dataset. Note that in Table III, Net I with all given network components, corresponding to JND-SalCAR, achieves the best performance in terms of prediction accuracy.

**Importance of the JND Probability Map**. Comparing Net A and I suggests that Jnd-Subnet plays a significant role in JND-SalCAR. The JND probability map provides subjective perceptual distortion information to the network and helps the network converge faster and make better score predictions.

**Design of SalCAR**. To demonstrate the superiority of SalCAR, four different block architectures were designed and tested. Fig. 5 shows the block designs and their details, with experimental results provided in Table III (Nets B, C, D, and I). In a comparison of Nets B and C, feeding SalCAR with the extracted features from the saliency maps improved the prediction performance, as doing so served as a form of spatial attention. Adding channel attention in Net D improved the prediction performance as well. Net D exhibited the largest performance drop among all combinations when a skip connection is eliminated from the original JND-SalCAR. The skip connection combines low-level features and high-level features after saliency-oriented spatial attention and channel attention.

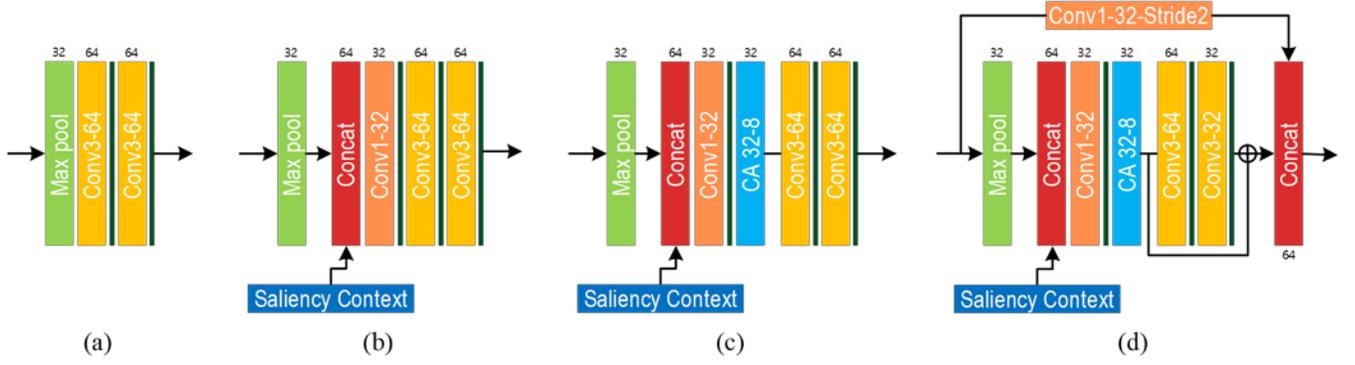

Fig. 5. Ablation study to find the effective structure of the SalCAR block: (a) is the baseline and (b) is (a) with a saliency context feature by concatenation. (c) is (b) with channel attention and (d) is (c) with two different types of skip connections, which results in the SalCAR block.

TABLE III. ABLATION STUDY ON DIFFERENT COMBINATIONS OF NETWORK COMPONENTS IN JND-SALCAR. PERFORMANCES ARE EVALUATED ON TID2013 DATASET.

| Net | JND Prob. Map | Saliency as Input | Channel Attention | Skip Connection | Rank Loss | Saliency-guided Loss | Split Convolution | SRCC |
|---|---|---|---|---|---|---|---|---|
| Baseline | ✗ | ✗ | ✗ | ✗ | ✗ | ✗ | ✗ | 0.937 |
| A | ✗ | ✓ | ✓ | ✓ | ✓ | ✓ | ✓ | 0.950 |
| B | ✓ | ✗ | ✗ | ✗ | ✓ | ✓ | ✓ | 0.943 |
| C | ✓ | ✓ | ✗ | ✗ | ✓ | ✓ | ✓ | 0.946 |
| D | ✓ | ✓ | ✓ | ✗ | ✓ | ✓ | ✓ | 0.948 |
| E | ✓ | ✓ | ✓ | ✓ | ✗ | ✗ | ✓ | 0.951 |
| F | ✓ | ✓ | ✓ | ✓ | ✓ | ✗ | ✓ | 0.955 |
| G | ✓ | ✓ | ✓ | ✓ | ✗ | ✓ | ✓ | 0.954 |
| H | ✓ | ✓ | ✓ | ✓ | ✓ | ✓ | ✗ | 0.957 |
| I | ✓ | ✓ | ✓ | ✓ | ✓ | ✓ | ✓ | 0.959 |

**Additional Losses**. Nets E, F and G are variations of JND-SalCAR in terms of loss functions. Both rank and saliency-guided losses contributed to the performance improvement of JND-SalCAR.

**Split Convolution**. Split convolution was designed to decrease the number of parameters involved in JND-SalCAR. As indicated in the parameter analysis results in Table IV, split convolution can greatly reduce the number of parameters while maintaining the performance of the network.

*D. Comparison of patch quality maps and patch weight maps*

We provide a detailed analysis with a comparison of patch weight and quality maps evaluated by the baseline network and JND-SalCAR. Five different types of distortions are used for the analysis: Gaussian blur (GB), JPEG compression (JPEG), JP2K compression (JP2K), additive white Gaussian noise (AWGN), and multiplicative Gaussian noise (MGN).

For GB, JPEG, and JP2K, the distortions in complex regions are more easily perceived than those in smooth regions. Therefore, complex regions should have lower values than smooth regions in the patch quality maps.

For AWGN and MGN, the distortions in smooth regions are much more noticeable than those in complex regions. Therefore, smooth regions should have lower values in patch quality maps.

In order to demonstrate the effectiveness of JND-SalCAR, we compare the patch quality maps and patch weight maps from two distorted images by GB and AWGN between JND-SalCAR and the baseline, as presented in Table III. Fig. 6 shows the patch quality and weight maps for the '*lighthouse*' image from the LIVE [3] dataset. The brighter regions indicate higher values in terms of quality and weight.

For the GB-distorted image shown in Fig. 6-(c), the distortions in complex regions (lighthouse) are mostly visible than the distortion in smooth regions (background). Therefore, complex regions should have lower values than smooth regions in the patch quality maps. The patch quality maps (Fig. 6-(l)) produced by JND-SalCAR are in good agreement with what was expected compared to those (Fig. 6-(k)) by the baseline.

For the AWGN-distorted image in Fig. 6-(e), the smooth region should have lower values in the patch quality maps. Both the baseline and JND-SalCAR presented results in good agreement with HVS. However, it should be noted that the smooth region clearly receives lower quality scores from the quality map evaluated by JND-SalCAR (Fig. 6-(n)) compared to the baseline.

TABLE IV. PARAMETER COMPARISONS. THE SRCC VALUES ARE BASED ON THE RESULTS FROM THE TID2013 DATASET.

| Methods | Bosse et al. [2] | PieAPP [4] | JND-SalCAR (non-split) | JND-SalCAR (with split) |
|---|---|---|---|---|
| Parameter Number | 6,262 K | 18,147 K | 6,404 K | **3,547** K |
| SRCC | 0.940 | 0.945 | 0.954 | **0.956** |

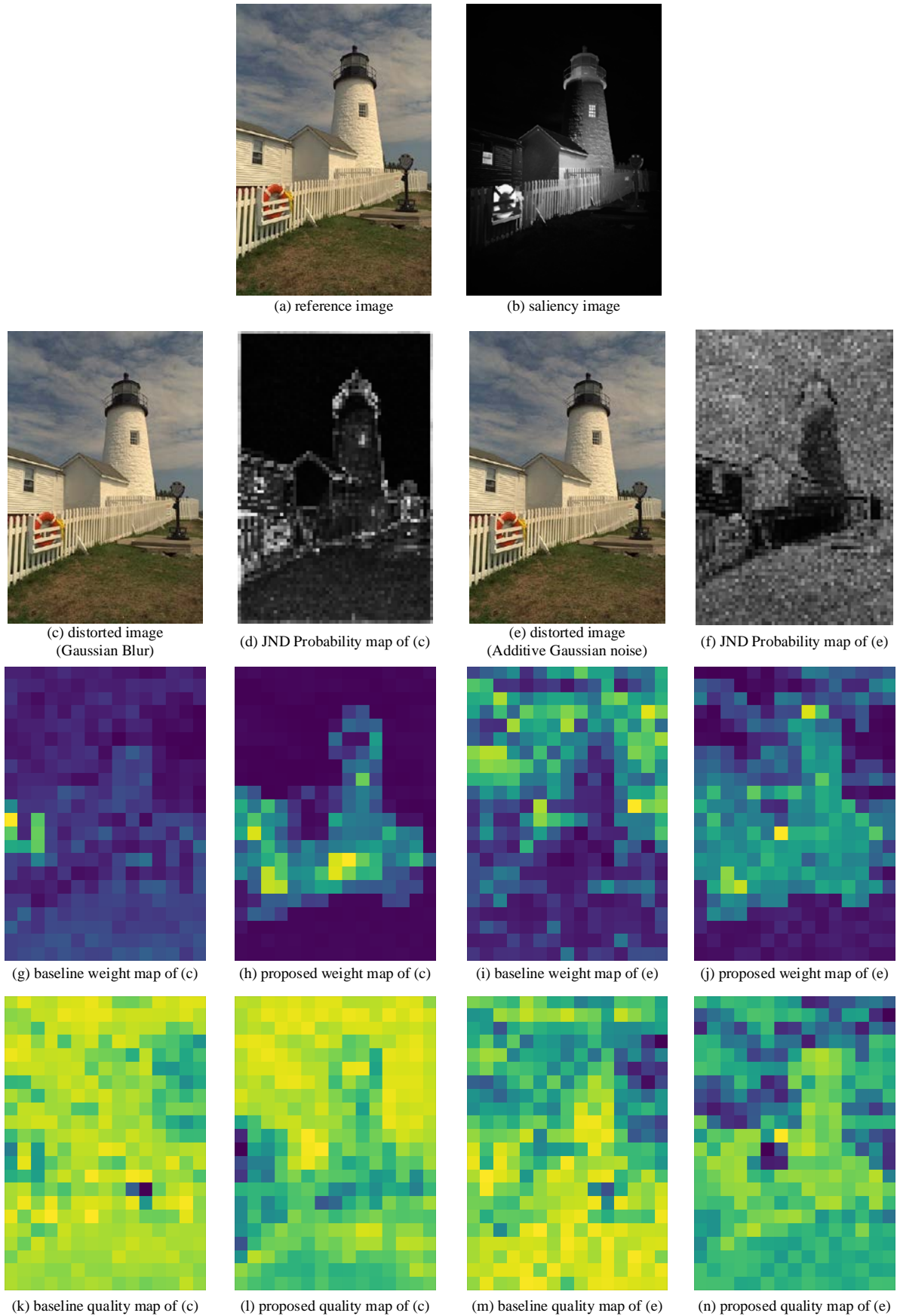

Fig. 6. Examples of patch quality maps and patch weight maps for images with different types of distortion for the same reference image ((a): '*lighthouse*'). For (c), the ground truth MOS value is 82.54 and the pMOS values are 91.11 by the baseline and 83.27 by JND-SalCAR. For (e), the ground truth MOS value is 81.83 and the pMOS values are 84.39 by the baseline and 79.99 by JND-SalCAR.

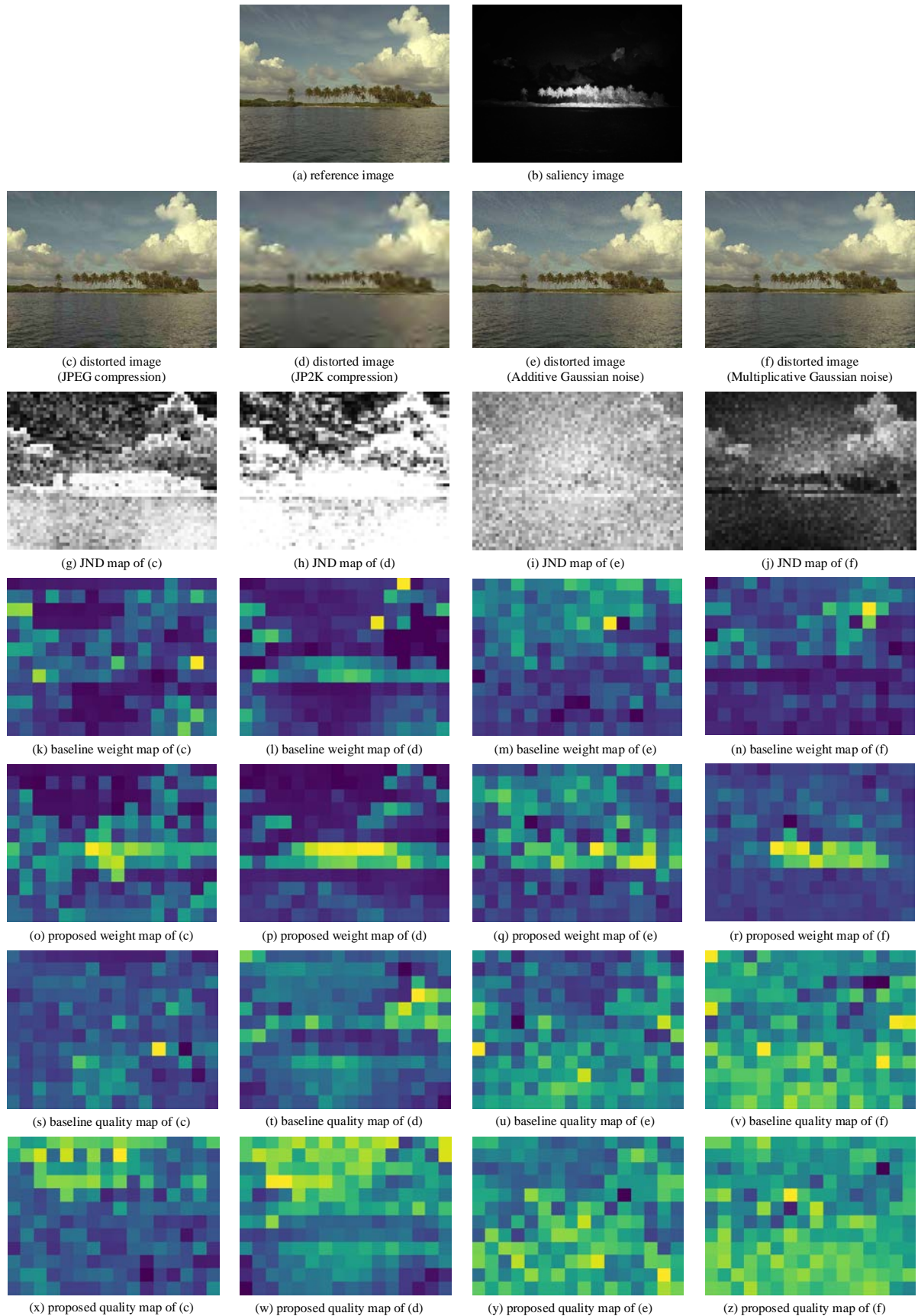

Fig. 7. Examples of patch quality maps and patch weight maps for images with different types of distortion for the same reference image (i16). For (c), the ground truth MOS value is 27.35 and the pMOS values are 23.63 by the baseline and 26.29 by JND-SalCAR. For (d), the ground truth MOS value is 10.56 and the pMOS values are 16.30 by the baseline and 15.25 by JND-SalCAR. For (e), the ground truth MOS value is 55.85 and the pMOS values are 53.71 by the baseline and 54.30 by JND-SalCAR. For (f), the ground truth MOS value is 61.94 and the pMOS values are 61.54 by the baseline and 62.06 by JND-SalCAR.

For patch weight maps, we incorporated a saliency map into the saliency loss term as a guideline for the patch weight maps. The patch weight maps (Fig. 6-(h), (j)) by JND-SalCAR successfully follow the saliency map guideline (Fig. 6-(b)), in good agreement with the HVS characteristic, while the baseline network completely fails to do so (Figs. 6-(c), (e)). Hence, we can infer that the saliency loss term effectively acts as a successful guideline for the learning of patch weight maps.

Fig. 7 shows another set of examples evaluated on GB, JP2K, AWGN, and MGN-distorted *'ocean'* image from the TID2013 [13] dataset, as produced by JND-SalCAR and the baseline. As noted earlier, the patch quality maps for JPEG and JP2K distortion should have lower values on textured areas (ocean). Figs. 7-(x), (w) clearly demonstrate this characteristic and show better results than the patch quality map evaluated by the baseline (Figs. 7-(s), (t)). The patch quality maps for AWGN and MGN distortions should have higher values in complex regions. The patch quality maps estimated by JND-SalCAR (Figs. 7-(y), (z)) successfully follow this HVS characteristic.

*E. Future works*

As for our future study, the JND-SalCAR can be extended to an NR-IQA network to predict the subjective quality scores of distorted images by incorporating a JND error map prediction network without reference. It can also be extended for other types of IQA, including quality assessments of dehazed images [76, 77], light field images [78], 3D images [79], virtual reality (VR) images [80], and audio-visual signals [81].

## V. CONCLUSION

In this paper, the proposed JND-SalCAR successfully reflects the human visual sensitivity and psychophysical characteristics in predictions of IQA problems by incorporating JND probability maps and saliency maps. The proposed SalCAR block clearly improved the prediction performance by extracting perceptually important features using saliency-based spatial attention and channel attention. Moreover, utilizing the visual saliency map as a guideline for predicting the patch weight map enables stable training of end-to-end optimization of the proposed JND-SalCAR. Through extensive experiments, JND-SalCAR was demonstrated to outperform state-of-the-art IQA methods on various IQA datasets, showing thus far the best performance in terms of SRCC and PLCC.

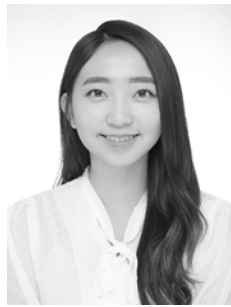
**Soomin Seo** received the B.S and M.S. degrees in the school of electrical engineering from Korea Advanced Institute of Science and Technology (KAIST), Daejeon, South Korea, in 2017 and 2019, respectively, where she is currently pursuing the Ph.D. degree. Her research interests include pan-sharpening, super-resolution, image restoration, image quality assessment, and deep learning.

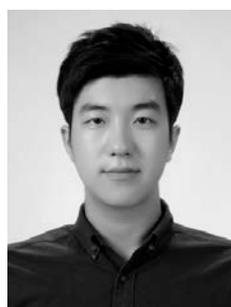
**Sehwan Ki** received the B.S. degree in electrical engineering from Kyungpook National University, Daegu, South Korea, in 2015, and the M.S. degree in electrical engineering from Korea Advanced Institute of Science and Technology (KAIST), Daejeon, South Korea, in 2017. He is currently pursuing the Ph.D. degree in electrical engineering at KAIST. His research interests include perceptual video coding, perceptual image/video quality assessment and deep-learning based image/video processing.

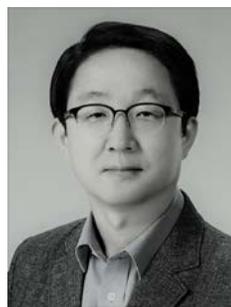
**Munchurl Kim** (M'99, SM'13) received the B.E. degree in Electronics from Kyungpook National University, Daegu, South Korea, in 1989, and the M.E. and Ph.D. degrees in Electrical and Computer Engineering from the University of Florida, Gainesville, in 1992 and 1996, respectively. After his graduation, he joined the Electronics and Telecommunications Research Institute, Daejeon, Korea, as a Senior Research Staff Member, where he led the Realistic Broadcasting Media Research Team. In 2001, he was an Assistant Professor with the School of Engineering, Information and Communications University (ICU), Daejeon. Since 2009, he has been with the School of Electrical Engineering, Korea Advanced Institute of Science and Technology (KAIST), Daejeon where he is now a Full Professor. He had been involved with scalable video coding and high efficiency video coding (HEVC) in JCT-VC standardization activities of ITU-T VCEG and ISO/IEC MPEG. His current research interests include deep learning for image restoration and visual quality enhancement, deep video compression, perceptual video coding, visual quality assessments, computational photography, machine learning and pattern recognition.